# Scrubbing Sensitive PHI Data from Medical Records made Easy by SpaCy - A Scalable Model Implementation Comparisons

Authors: Rashmi Jain, Dinah Samuel Anand, Vijayalakshmi Janakiraman
{rashmij, dinahs, vijayalakshmij}@gradvalley.in
GradValley Translational Data Science Research Lab, Coimbatore, India

## Abstract

De-identification of clinical records is extremely important process which enables the use of the wealth of information present in them. There are a lot of techniques available for this but none of the method implementation has evaluated the scalability, which is an important benchmark. We evaluated numerous deep learning techniques such as BiLSTM-CNN, IDCNN, CRF, BiLSTM-CRF, SpaCy, etc. on both the performance and efficiency. We propose that the SpaCy model implementation for scrubbing sensitive PHI data from medical records is both well performing and extremely efficient compared to other published models.

## Introduction

Clinical records contain an immense amount of information which is highly beneficial for clinical research. However, what it also contains is patient's personal information which is considered as protected health information (PHI) based on Health Insurance Portability and Accountability Act of 1996 (HIPAA) Privacy Rule, restricting its usage.

Privacy of patient's information is a crucial part of any Electronic Health Record System and any kind of leak of these personal identifiers is a serious breach of privacy. Therefore, there is a need of de-identification of clinical records, so that it can be used in its full potential for clinical research. Also, when a patient is aware that their health information is meticulously protected, there is a higher chance of transparency and correctness from the provider regarding certain specific health concerns.

However, the de-identification of clinical records is often carried out manually which is both a resource intensive and a time consuming task. Dorr et al [1] have assessed that manually de-identifying narrative text notes takes an average time of 87.2 ± 61 seconds per note, and concluded that it is a tedious and time-consuming process and also difficult to exclude all PHIs required by HIPAA. Hence, automatic de-identifying of clinical records is needed in such a situation.

There are majorly two different approaches used for de-identification: rule based/pattern matching and machine learning. Some of the techniques uses a combination of both pattern matching and machine learning. According to Meystre et al [10], algorithms in machine learning used for de-identification of clinical records includes Conditional Random Fields(CRF) [3][4][5], Support Vector Machines (SVM) [6][7][8], Decision Trees [9], etc. Most of these techniques except Gardner et.al [4] use some kind of knowledge source or semantic information along with the machine learning algorithm. Usage of medical dictionaries makes it easier to identify PHIs in the record but it also makes it difficult to scale the approach for industrial usage by hospitals because of the lack of efficiency specifically, in terms of time.

With the advent of state of the art deep learning algorithms [11][12][13][14], various researchers have tried to implement it with the problem statement at hand. Particularly, an interesting approach is used in Dilip et al [13], where the team uses a combination of ID-CNN, biLSTM CNN and regular expressions to scrub medical records. As the model produced large number of false positives, they used BiLSTM CNN only for identifying names and added an iterated dilated convolutions model to improve its performance. Both the models were trained separately on Ontonotes dataset where Bi-LSTM detected "NAME" entity while ID-CNN model captured "NAME", "ADDRESS" and "DATE" entities. They evaluated both the model's performance on Ontonotes dataset. The ID-CNN model achieved a segmented micro F1 score of 86.84 while Bi-LSTM model achieves 86.5. Table 1 shows the comparative analysis of various methods and techniques handled by various authors.

| Author | Year of Publication | ML Algorithm | Technique/Method |
|---|---|---|---|
| Aramaki et al [3] | 2006 | CRF | 2 step learning process, first, learning with local features (such as the surrounding words, POS tags, casing, etc), sentence features (such as sentence position, length of surrounding sentences, etc) and extra resources features (such as the presence of the person name, location, and date in a predefined dictionary), second, learning using additional features from the first learning results. |
| Guo et al [6] | 2006 | SVM | Trained SVM classifiers for PHI entity types by combining basic token level information. existing ANNIE (A Nearly-New Information Extraction system) and newly created JAPE grammar rules along with augmented lists. |
| Szarvas et al [9] | 2007 | Decision Trees | Used ML to classify Named Entities in newswire articles, they used two new features, regular expressions for the well-defined classes and subject heading information and introduced an iterative learning approach |
| Wellner et al [5] | 2007 | CRF/HMM | Considered de-id as a sequence labeling task, used two existing toolkits called Carafe (implementing CRF) and LingPipe (Hidden Markov Models). The best performing system came out to be the one using Carafe. |
| Gardner et al [4] | 2008 | CRF | Their tagging process involved initial tagging of a small set of reports, automatic tagging for the rest of the reports with our attribute extraction component using the small training set, and manual retagging or correction for all the reports. This dataset was used for NER using CRF for extracting identifying and sensitive attributes. |
| Uzuner et al [8] | 2008 | SVM | De-identified medical discharge summaries based on SVM and local context (features of the target and its close neighbours). In addition to this, they use Link Grammar Parser to obtain syntactic information as well as medical dictionaries such as MeSH of UMLS. |
| Shweta et al [12] | 2016 | RNN | They used deep neural network based approach for patient de-identification. They implemented and compared different variants of RNN architecture, including Elman and Jordan. |
| Dernoncourt et al [11] | 2017 | ANN | It is the first de-identification system which is based on artificial neural networks (ANNs), and it requires no handcrafted features or rules |

| Khin et al [14] | 2018 | BiLSTM-CRF | Used BiLSTM-CRF method with a1394 dimensional word embedding (concatenating the GloVe and ELMo with the character embedding vector, POS one-hot-encoded vector, and the casing embedded vector) to deidentify medical records. |

**Table 1: Summary of techniques and methods used by various authors to solve de-identification task**

Deep learning techniques in general ([11][12][13][14]) have proved to be of high accuracy for de-identification tasks such as, an F1 score of 99% on MIMIC discharge summaries in Dernoncourt et al. But none of the work so far evaluated the scalability, which we have identified as an important benchmark for the task.

## Method

### Data

We have used de-identification challenge data set by i2b2 provided in their 2014 De-identification and Heart Disease Risk Factors Challenge[2]. The dataset was annotated and all the PHIs were given an XML tag indicating its category and type, wherever applicable. 1173 out of 1304 files were used for training and remaining 131 were used for testing (a train-test split ratio of 9:1).

| PHI category | Sub-category |
|---|---|
| NAME | PATIENT, DOCTOR, USERNAME |
| PROFESSION | (none) |
| LOCATION | HOSPITAL, ORGANIZATION, STREET, CITY, STATE, COUNTRY, ZIP, OTHER |
| AGE | (none) |
| DATE | (none) |
| CONTACT | PHONE, FAX, EMAIL, URL, IPADDRESS |
| IDs | SOCIAL SECURITY NUMBER, MEDICAL RECORD NUMBER, HEALTH PLAN NUMBER, ACCOUNT NUMBER, LICENSE NUMBER, VEHICLE ID, DEVICE ID, BIOMETRIC ID, ID NUMBER |

**Fig 1: i2b2 annotation categories and sub-categories**

We made a few changes in the tags mentioned above such as -
1. Combining sub categories for NAME PHI category under one tag called NAME
2. Combining Hospital and Organization subcategory under one tag called ORG

PD Scrubber is a scalable tool that automatically identifies all PHIs according to HIPAA guidelines and scrubs them from a given clinical record. We have used the machine learning technique and no knowledge resource or pre-built dictionaries is used to identify PHI from a medical record. This makes the tool scalable and extremely efficient

We have also used Ontonotes corpus [15] for trying the modified approach from Dilip et al [13].

### Models

Machine learning models were used to identify names, organization, street, city, state, country, zip, age, and date. We tried multiple machine learning models and techniques to solve this task:

1. IDCNN + BiLSTM + Regex for entity identification and RNN + UMLS Metathesaurus API[16] for Medical Term Disambiguation(MTD) on Ontonotes corpus
2. CRF and BiLSTM CRF for i2b2 corpus
3. Dilated CNN and BiLSTM CNN for de-identification on i2b2 corpus
4. spaCy for de-identification on i2b2 corpus

**Approach 1: IDCNN + BiLSTM + Regex for entity identification and RNN + UMLS Metathesaurus API for MTD on Ontonotes corpus**

We started experimenting the approach presented by Dilip et al [13] with a few modifications in the architecture such as

1. Architecture of the IDCNN model
2. MTD
3. Stacking of the models

We used IDCNN model from Emma et al., [17] because of the higher F1 scores than Dilip et al [13]. MTD modifications and stacking of all the models were carried out because there was very little information available in the published article. Architecture for the modified approach is given in Fig 2. In this approach, we used a combination of IDCNN, BiLSTM CNN and regular expressions for identifying PHIs.

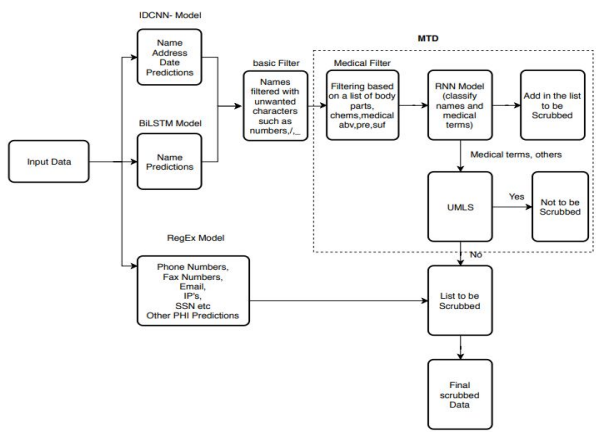

**Fig 2: Scrubbing architecture for Approach 1**

First step of this approach was to detect all the entities using BiLSTM, IDCNN and regex models. Since regex model was based on fixed patterns there is no need of MTD and these entities were directly scrubbed from the medical text. From the other two models, the entities were passed through a basic filter which removed the names with unwanted characters such as underscores, slashes, hashes, etc.

Second step of this approach was MTD. Since the training is done on a generic Ontonotes corpus, several terms are misclassified by the models. Dilip et al [13], used Fuzzy approach over a dictionary to identify medical terms. We tried a combination of dictionaries (names of body parts, medical abbreviations, chemicals, medical prefixes and suffixes), RNN model, and UMLS Metathesaurus API.

The remaining entities from the basic filter were passed through the medical filter to remove a few obvious medical terms such as names of body parts, chemicals, abbreviations, etc. We also trained an RNN model to classify between names and medical terms trained on a list of medical terms and names. A character level embedding was used for RNN model. The model achieved an accuracy of 80%. We did a double check on the entities identified as medical terms, using UMLS Metathesaurus API because it is important to remove all the PHIs, especially names from the medical record. List of names from RNN and the list of non medical terms from UMLS were added to the entities to be scrubbed.

*Model Outcome and Challenge*

Scrubbing task from this method gave good results as in Fig 3 but the performance (time taken) was extremely huge and the approach was also non-scalable. We performed scrubbing on a medical record of 650 KB, and IDCNN could not predict the output and kept on killing the server.

12 years ago Mr. PERSON was previously seen on DATE for evaluation of chronic GE reflux disease. Subsequently, an EGD was performed on DATE concluding a small hiatal hernia and gastritis from which biopsies were obtained. 15 years ago there is no evidence of peptic ulcer disease or neoplasm. Pathology findings document no evidence of Barrett's esophagus or PERSON Pylori infection. His phone number is NUMBERS and email id is EMAIL His policy number is NUMBERS

**Fig 3: Scrubbing output by Approach**

We identified that the most time consuming and challenging piece of this approach is MTD and in order to achieve a scalable approach we have to either remove or bypass the step without affecting the model's performance. Hence, it was decided to achieve scrubbing through a medical corpus which was i2b2 2014 De-identification challenge dataset.

**CRF (Approach 2) and BiLSTM-CRF (Approach 3) for de-identification on i2b2 corpus**

We tried a simple Conditional Random Field (CRF) model with the feature set as the casing of the word and the POS tags of the word in the sentence. The model was trained using 'lbfgs' algorithm and C1 =0.1 and C2 =1e-3 for a maximum iterations of 100.

We also tried a BiLSTM CRF model with the bag of words 200 dimensional embedding. A single BiLSTM layer was used consisting of 200 units and a dropout of 0.25. Then a time distributed dense layer with 'relu' activation function was used followed by a CRF layer. The optimizer used for training is 'rmsprop'.

**Dilated CNN (Approach 4) and BiLSTM CNN (Approach 5) for de-identification on i2b2 corpus**

ID-CNN from Approach 1 was one of the reasons of the non-scalability of the product. Hence we developed and experimented a new model, Dilated CNN with an alternative preprocessing technique. The word embedding used for the Dilated CNN and Bi-LSTM-CNN was Glove 840 billion words and 300 dimensional vector. We assumed that we do not require MTD as we are using i2b2 2014 De-identification challenge dataset.

The results were equally good for both the models. As assumed, we didn't require any MTD and now that the model was trained on medical corpus. The Dilated CNN didn't break on the 2MB file but the time taken for both the models were extremely high. It ran more than a day for a complete prediction and for this reason alone, the main objective of having a scalable product with efficiency and performance may not be achieved with this approach.

**Approach 6 and Approach 7: spaCy for de-identification on i2b2 corpus**

spaCy - an industrial strength natural processing in python has neural models for tagging, parsing and entity recognition. It is an open source free library for advanced NLP. We used spaCy v 2.1 for our named entity recognition task. Two different techniques were used to train the spaCy model. First, every medical record was used as one training sentence along with the entities and the spans were calculated considering the same. Second, every sentence in the medical record was used as one training sentence. For the first technique, the i2b2 dataset was converted into spacy format. 100 epochs and a drop of 0.5 was used for training (**spacy 1**) (**Approach 6**).

For the second technique, all the records were sentence tokenized using NLTK's Punkt tokenizer with few added salutations such as Mr., Ms., Dr., etc. Model was trained for 60 epochs and a drop of 0.25 (**spacy 2**) (**Approach 7**).

**Regular expressions (regex)**

A regex model was developed for identifying the entities that have a fixed pattern such as Email Id, URLs, IP Addresses, SSN, Phone numbers and other IDs.

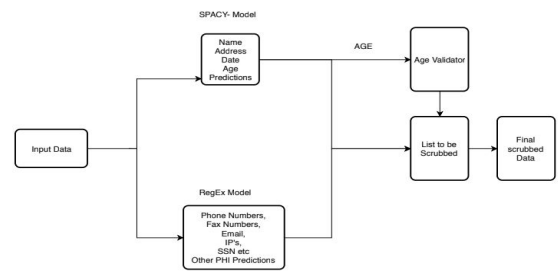

**Fig 4: Final Scrubber Architecture**

12 years ago Mr. PERSON was previously seen on DATE for evaluation of chronic GE reflux disease. Subsequently, an EGD was performed on DATE concluding a small hiatal hernia and gastritis from which biopsies were obtained. 15 years ago there is no evidence of peptic ulcer disease or neoplasm. Pathology findings document no evidence of Barrett's esophagus or H. Pylori infection. His phone number is NUMBERS and email id is EMAIL. His policy number is NUMBERS

**Fig 5: Final scrubber Output**

**Results**

F1 score was used as a metric for evaluating the model performance on the i2b2 testing dataset of 131 medical records. Also, time cost was tested on a 2 MB file that was collated using multiple files from i2b2 test dataset. For the approach on ontonotes corpus, F1 scores for BiLSTM CNN and IDCNN are 81.9% and 84.5% respectively. The IDCNN model could not scrub a 2 MB file, killing the whole system. On reducing the size, it broke even at a file size of 650 KB.

i2b2-Dilated CNN and i2b2-BiLSTM-CNN achieved an F1 score of 87.1% and 91.8% respectively. Time taken by both i2b2-Dilated CNN and i2b2-BiLSTM-CNN for prediction on the 2 MB file was **more than a day**.

spaCy 1 and spacy 2 achieved an F1 score of 80.2% and 91.2% respectively. spaCy 1 could not handle a 2 MB file with a value error saying "Text exceeds maximum of 1000000" whereas spaCy 2 took around **3.3 minutes** for the prediction.

CRF and BiLSTM CRF achieved an F1 score of 76.7% and 77% resp. With such a low F1 score as compared to other models, we did not consider these models for time cost calculation.

Combined result is shown in Table

| Approach Name | Model Name | F1 score | Time taken | Reason for picking up the model | Cons |
|---|---|---|---|---|---|
| Modified Dilip et al, Approach 1 | Ontonotes IDCNN | 84.50% | - | Emma's implementation [17] | Could not predict for sample file, broke the system |
| | Ontonotes BiLSTM CNN | 81.90% | - | Dilip et al | |
| Approach 2 | i2b2-CRF | 76.70% | - | [3] suggested the use of CRF for i2b2 de-id challenge, achieved 94% f1 score | Could not achieve a good accuracy with the feature vectors |
| Approach 3 | i2b2-BiLSTM-CRF | 77% | 83 mins | Proven model for NER on Ontonotes corpus [18] | Time as well as F1 measure, we could have hypertuned, but then time taken for a very simple embeddings was huge, change in embeddings could have increased the time taken for prediction |
| Approach 4 | i2b2-Dilated CNN | 87.10% | > 1440 mins | Modification from Emma's implementation to see if it will break or not | Time taken for prediction on a 2 MB test file was more than 1 day |

| Approach 5 | i2b2-BiLSTM-CNN | 91.80% | > 1440 mins | Models developed from modified Dilip et al with training on i2b2 dataset instead of Ontonotes corpus | Time taken for prediction on a 2 MB test file was more than 1 day |
| Approach 6 | spaCy 1 | 80.20% | - | Sentence tokenization is not perfect when considering an unstructured text, so tried the training without sentence tokenization | It did not predict the output for 2 MB file because of the max size limit, and also the F1 score was low |
| Approach 7 | spaCy 2 | 91.20% | ~3.3 mins | Proven tech for NER which is also industrial in nature, hence scalable | F1 score could still be improved, age prediction is not very accurate, could be improved further |

**Table 2: Combined results for all the experimented approaches for de-identification of medical records**

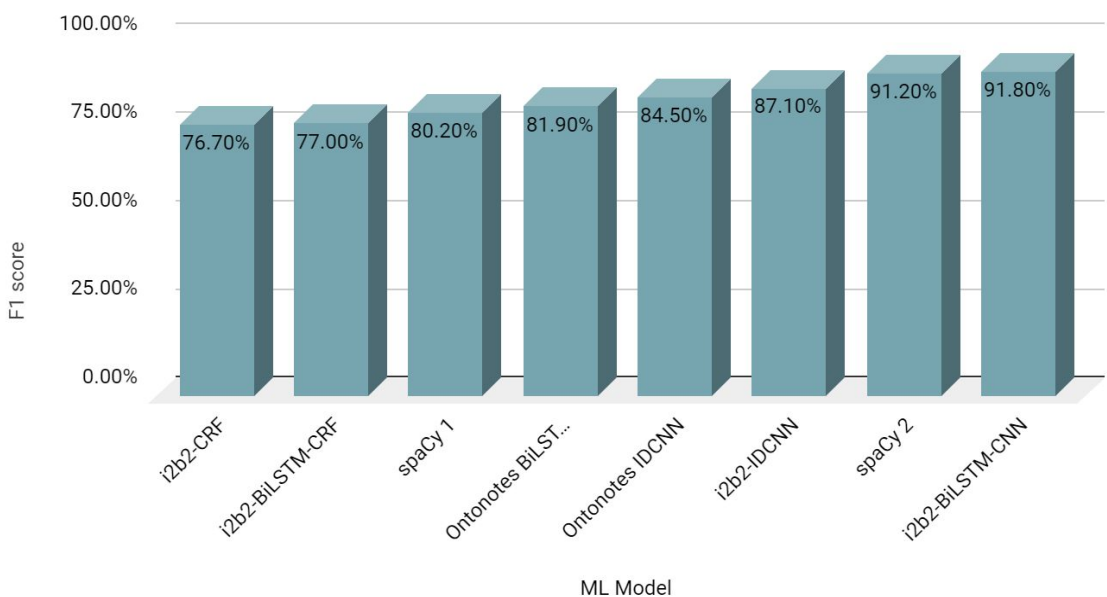

**Fig 6: Chart showing a comparison of F1 scores of various ML Models**

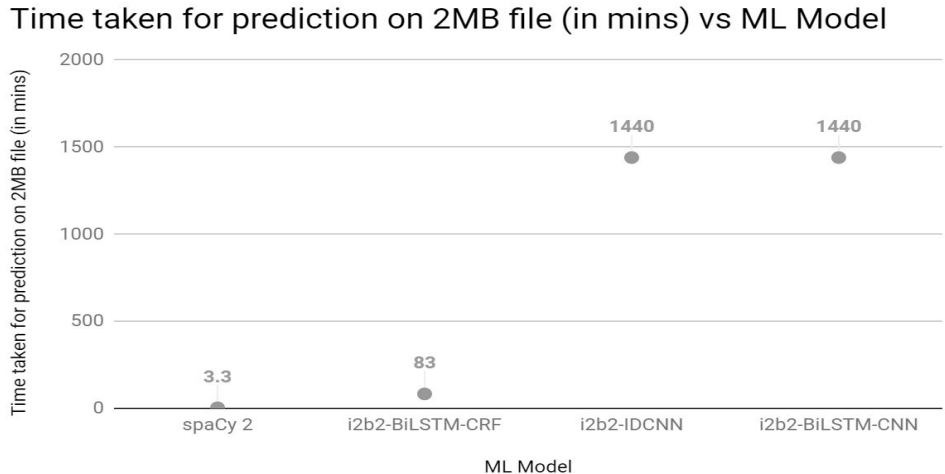

**Fig 7: Chart showing a comparison of time taken for prediction of various ML Models**

## Conclusions

Major challenge that we faced with Approach 1, was the performance (time taken) and scalability. We also realised that MTD took a considerable amount of time which results in poor performance. Therefore, domain specific tasks should use datasets that are highly contextual and related to the domain in order to save time and hence a shift from Ontonotes corpus to i2b2 dataset.

i2b2-BiLSTM-CNN was the best model in terms of F1 score, whereas spaCy 2 model proved to be the best in terms of time cost and second best in terms of F1 score. Since, one of our important benchmarks is the time cost, we propose spaCy 2 as the best model to scrub PHI information from the medical records.

This is one of the first works presenting the usage of spaCy for de-identification of clinical records.

There are multiple products available online to scrub medical data, however most of them are downloadable applications and none of them is a web application. To make the scrubber more accessible to the community, we have developed a web application for the same which can be accessed via the link (http://pdscrubber.gradvalley.in:83/). An API for developers is available for developers to use it as a part of their applications.

## Recommendations

Personal Data scrubbing is a task that can be applied to pretty much any domain such as retail, finances, ecommerce, etc. The approach presented in the paper is highly specific to medical domain because of the training dataset, we wish to expand it to multiple domains. Diverse training dataset can be added to improve the F1 measure of the model.

Also, we have currently explored CRF and BiLSTM CRF only with simple feature vectors, we can improve the individual F1 scores by exploring more types of word vectors and embeddings.

While using regex, we have combined most of the entities as IDs and Numbers for phone numbers, where we expect the model to be more exhaustive and be able to identify individual IDs.

We have developed a web application that currently support only text files to be scrubbed through the web application, that needs to be scaled for other document formats.

## Acknowledgement

The authors would like to thank Ms Sudha V for Project Management support and Mr Raja S for the web application development and deployment. Deidentified clinical records used in this research were provided by the i2b2 National Center for Biomedical Computing funded by U54LM008748 and were originally prepared for the Shared Tasks for Challenges in NLP for Clinical Data organized by Dr. Ozlem Uzuner, i2b2 and SUNY.

## References

[1] A Dorr, D & F Phillips, W & Phansalkar, Shobha & A Sims, S & F Hurdle, J. Assessing the Difficulty and Time Cost of De-identification in Clinical Narratives. Methods of information in medicine. 2006. 45 246-52.

[2] Stubbs A, Uzuner Ö. Annotating longitudinal clinical narratives for de-identification: The 2014 i2b2/UTHealth corpus. J Biomed Inform. 2015 Dec; 58(Suppl): S20–S29.

[3] Özlem Uzuner, Yuan Luo, Peter Szolovits, Evaluating the State-of-the-Art in Automatic De-identification, J Am Med Inform Assoc. 2007 Sep-Oct;14(5):550-63.

[4] Gardner J, Xiong L. HIDE: An Integrated System for Health Information De-identification. Proceedings of the 2008 21st IEEE International Symposium on Computer-Based Medical Systems. 2008. pp. 254–9.

[5] Wellner B. Rapidly retargetable approaches to de-identification in medical records. J Am Med Inform Assoc. 2007;14:564–573

[6] Guo Y. Identifying Personal Health Information Using Support Vector Machines. i2b2 Workshop on Challenges in Natural Language Processing for Clinical Data, Washington, DC. 2006

[7] Hara K. Applying a SVM Based Chunker and a Text Classifier to the Deid Challenge. i2b2 Workshop on Challenges in Natural Language Processing for Clinical Data, Washington, DC. 2006.

[8] Uzuner O. A de-identifier for medical discharge summaries. Artif Intell Med. 2008;42(1):13–35.

[9] Szarvas G, Farkas R, Busa-Fekete R. State-of-the-art anonymization of medical records using an iterative machine learning framework. J Am Med Inform Assoc. 2007. pp. 574–80.

[10] Meystre SM, Friedlin FJ, South BR, Shen S, Samore MH. Automatic de-identification of textual documents in the electronic health record: a review of recent research. BMC Med Res Methodol. 2010 Aug 2;10:70.

[11] Dernoncourt, Franck & Young Lee, Ji & Uzuner, Ozlem & Szolovits, Peter. De-identification of Patient Notes with Recurrent Neural Networks. Journal of the American Medical Informatics Association : JAMIA. 2016. May 1;24(3):596-606.

[12] Yadav, S., Ekbal, A., Saha, S., & Bhattacharyya, P. Deep Learning Architecture for Patient Data De-identification in Clinical Records. Proceedings of the Clinical Natural Language Processing Workshop, pages 32–41, Osaka, Japan. 2016. December 11-17.

[13]: Dilip, A.K., KamalRaj, K., & Sankarasubbu, M. PHI Scrubber: A Deep Learning Approach. 2018. arxiv.1808.0118

[14]: Khin, K., Burckhardt, P., & Padman, R. A Deep Learning Architecture for De-identification of Patient Notes: Implementation and Evaluation. 2018. arXiv:1810.01570v1

[15]: Eduard Hovy, Mitchell Marcus, Martha Palmer, Lance Ramshaw, and Ralph Weischedel. Ontonotes: The 90 In Proceedings of the Human Language Technology Conference of the NAACL, Companion Volume: Short Papers, NAACL-Short '06, pages 57–60, Stroudsburg, PA, USA, 2006. Association for Computational Linguistics.

[16]: Bodenreider O. The Unified Medical Language System (UMLS): integrating biomedical terminology. Nucleic Acids Res. 2004 Jan 1;32:D267-70.

[17]: Strubell, Emma & Verga, Patrick & Belanger, David & Mccallum, Andrew. Fast and Accurate Entity Recognition with Iterated Dilated Convolutions. 2017. (arXiv:1702.02098v3) 2670-2680.

[18]: Ghaddar, A., & Langlais, P. Robust Lexical Features for Improved Neural Network Named-Entity Recognition. Proceedings of the 27th International Conference on Computational Linguistics, pages 1896–1907 Santa Fe, New Mexico, USA. 2018 August 20-26.